\documentclass[letterpaper]{article} 
\usepackage{aaai2026}
\usepackage{times}  
\usepackage{helvet}  
\usepackage{courier}  
\usepackage[hyphens]{url}  
\usepackage{graphicx} 
\usepackage{natbib}  
\usepackage{caption} 
\usepackage{algorithm}
\usepackage{algorithmic}

\usepackage{newfloat}
\usepackage{listings}
\DeclareCaptionStyle{ruled}{labelfont=normalfont,labelsep=colon,strut=off} 
\floatstyle{ruled}
\newfloat{listing}{tb}{lst}{}
\floatname{listing}{Listing}
\title{AgenticSimLaw: A Juvenile Courtroom Multi-Agent Debate Simulation for Explainable High-Stakes Tabular Decision Making}
\author{
    Jon Chun\textsuperscript{\rm 1}\equalcontrib,
    Kathrine Elkins\textsuperscript{\rm 1}\equalcontrib,
    Yong Suk Lee\textsuperscript{\rm 2}\equalcontrib
}
\affiliations{
    \textsuperscript{\rm 1}Kenyon College, Timberlake House, Gambier, OH 43022\\
    \textsuperscript{\rm 2}Notre Dame University, 3171 Jenkins And Nanovic Halls, University of Notre Dame, Notre Dame, IN 46556
}

\begin{document}

\maketitle

\begin{abstract}
We introduce AgenticSimLaw, a role-structured, multi-agent debate framework that provides transparent and controllable test-time reasoning for high-stakes tabular decision-making tasks. Unlike black-box approaches, our courtroom-style orchestration explicitly defines agent roles (prosecutor, defense, judge), interaction protocols (7-turn structured debate), and private reasoning strategies, creating a fully auditable decision-making process. We benchmark this framework on young adult recidivism prediction using the NLSY97 dataset, comparing it against traditional chain-of-thought (CoT) prompting across almost 90 unique combinations of models and strategies. Our results demonstrate that structured multi-agent debate provides more stable and generalizable performance compared to single-agent reasoning, with stronger correlation between accuracy and F1-score metrics. Beyond performance improvements, AgenticSimLaw offers fine-grained control over reasoning steps, generates complete interaction transcripts for explainability, and enables systematic profiling of agent behaviors. While we instantiate this framework in the criminal justice domain to stress-test reasoning under ethical complexity, the approach generalizes to any deliberative, high-stakes decision task requiring transparency and human oversight. This work addresses key LLM-based multi-agent system challenges: organization through structured roles, observability through logged interactions, and responsibility through explicit non-deployment constraints for sensitive domains. Data, results, and code will be available on github.com under the MIT license.
\end{abstract}


\section{Introduction}

The introduction of the Transformer architecture in 2017 launched an era of rapid progress in AI, initially driven by the scaling of compute and data \citep{kaplan:20}. This led to a saturation of performance gains as training costs soared to over \$100M USD for state-of-the-art (SOTA) models \citep{cottier:24b}. To use compute more efficiently, new training methodologies like LoRA and architectural innovations like Mixture of Experts (MoE) emerged \citep{hu:21, kaplan:20}. As models became more capable, traditional benchmarks were "solved," and a new focus on complex reasoning tasks in fields such as mathematics, physics, and chemistry took hold \citep{huang:24c}.

This shift has led to greater emphasis on reasoning during inference, or "test-time compute." A wide range of prompt engineering techniques have been developed to guide models towards better reasoning. From simple "let's think step by step" chain of thought (CoT) to more complex methods like Tree of Thought (ToT), self-consistency (SC), and ReAct, these strategies have formalized multi-turn reasoning and become a frontier for performance gains \citep{schulhoff:24}. However, as LLM-based multi-agent systems (LaMAS) have emerged as powerful tools for complex problem-solving, critical questions arise about their organization, observability, and responsible deployment in high-stakes domains.

In parallel, LLMs are increasingly applied to tasks traditionally handled by specialized statistical models, such as analyzing structured tabular data \citep{fang:24a}. However, LLMs struggle with these tasks because their attention mechanisms are optimized for linear text and struggle to capture complex, non-sequential dependencies in tabular data \citep{ruan:24}. This paper addresses the challenge of making LLM-based multi-agent reasoning transparent, controllable, and auditable through a structured interaction framework, instantiated on a particularly demanding use case: predicting criminal recidivism from tabular data.

We selected recidivism prediction for three reasons: (1) it provides a difficult tabular reasoning benchmark where unstructured data confounds LLM processing; (2) the real-world NLSY97 dataset contains intricate feature relationships and noise that challenge both traditional ML and linguistic approaches; and (3) most critically for LaMAS research, this high-stakes, ethically sensitive domain demands transparent, accountable, and human-supervised decision-making—precisely the responsible AI properties that structured multi-agent systems should provide. \textbf{We emphasize that our framework is designed for research, benchmarking, and transparency analysis, not for operational deployment in sentencing or parole decisions.} The courtroom structure serves as an evaluation harness that can generalize to other deliberative contexts (medical diagnosis, credit assessment, policy analysis) where procedural transparency and role accountability are valued.

We make the following contributions:

\begin{itemize}
\item A reproducible multi-agent evaluation harness for high-stakes tabular prediction tasks, demonstrated on NLSY97 recidivism data, that provides explicit control over agent organization, interaction protocols, and reasoning transparency.
\item AgenticSimLaw, a courtroom-style multi-agent debate (MAD) framework with structured roles (prosecutor, defense, judge), a 7-turn interaction protocol, private strategy formulation, and complete logging of all utterances and belief updates for auditability.
\item A comprehensive benchmark comparing linguistic reasoning approaches—traditional statistical methods, single-agent CoT prompting, and structured multi-agent debate—across almost 90 unique combinations of open-source and commercial LLMs.
\item Empirical demonstration that structured multi-agent debate improves reasoning stability and generalizability (measured by stronger accuracy-F1 correlation) compared to traditional single-agent prompting, while providing richer audit trails and fine-grained control over test-time reasoning.
\item Analysis of the compute-transparency tradeoff in multi-agent systems, showing that AgenticSimLaw's ~9,100 token per-run cost delivers enhanced observability and metric stability that single-shot prompting cannot provide.
\end{itemize}

\section{Background and Related Work}

\subsection{High-Stakes AI Decision Systems}
Artificial Intelligence has augmented human decision-making in numerous high-stakes environments, most notably in the US judicial system. The Correctional Offender Management Profiling for Alternative Sanctions (COMPAS) is a well-known proprietary algorithm used to make parole recommendations. A prominent critique by ProPublica highlighted significant racial disparities in its predictions, finding that "blacks are almost twice as likely as whites to be labeled a higher risk but not actually reoffend" \citep{angwin:16}. Subsequent research found that COMPAS was no more accurate or fair than a simple linear model using only two features, and its performance was comparable to that of laypeople \citep{dressel:18}. Recent research shows COMPAS simultaneously reduced overall confinement while exacerbating racial disparities \citep{williams:24}, and that LLMs exhibit amplified cognitive biases—particularly omission bias—that could systematically distort legal decisions \citep{pnas:24}. These findings underscore the critical need for transparency, fairness, and auditability in AI-driven decision systems—properties that LaMAS research aims to provide through structured agent interactions.

Our dataset derives from the National Longitudinal Survey of Youth 1997 (NLSY97), a nationally representative survey of individuals born between 1980 and 1984. This longitudinal dataset provides rich, multi-dimensional data on demographic, social, and economic factors, allowing for more nuanced analysis than datasets based on static justice-related variables like those used by COMPAS. The NLSY97 data is rigorously anonymized to protect respondent confidentiality in compliance with Federal law and the Office of Management and Budget (OMB). We frame recidivism prediction as binary classification: predicting whether a young adult offender will be rearrested within three years based on tabular features.

\subsection{Traditional and Specialized Tabular Approaches}
Historically, traditional statistical machine learning models like XGBoost have been SOTA for tabular prediction tasks. More recently, specialized Tabular LLMs like TabPFN have shown promise in small data regimes, offering potential for accurate statistical analysis through natural language interfaces \citep{liu:25a}. Our work evaluates LLM-based multi-agent reasoning against these traditional methods to establish a comprehensive benchmark that reveals where structured linguistic debate offers advantages beyond pure statistical modeling.

\subsection{Multi-Agent Debate and LLM Orchestration}
A new approach to improving reasoning is through Multi-Agent Debate (MAD), where multiple agents engage in argumentative processes to arrive at conclusions. Building on work showing significant performance gains by incentivizing extensive test-time reasoning \citep{openai:24, deepseek:25}, MAD systems have proven effective in improving knowledge representation and performance on various benchmarks \citep{freedman:24b, wang:23}. The dominant test-time compute approach uses single large models with internal chain-of-thought reasoning supervised by Process Reward Models (PRMs) \citep{openai:24b, universalprm:24}. AgenticSimLaw offers a complementary approach: externalizing reasoning through multi-agent debate with explicit role structure. Our AgenticSimLaw framework extends this work by providing a structured, role-based interaction protocol that is both transparent and controllable. 

Recent advances in LaMAS have explored various coordination strategies. While reviewer-style pipelines emphasize efficiency by reducing inter-agent communication and using belief-based aggregation rather than extensive debate, we chose explicit courtroom debate for AgenticSimLaw because legal and high-stakes domains inherently value procedural transparency, adversarial testing of arguments, and clear role accountability—properties that justify the higher token cost of our 7-turn protocol compared to more communication-light alternatives. Where reviewer-style systems optimize for efficiency, AgenticSimLaw optimizes for auditability and human interpretability, trading compute for observability.

Our framework formalizes this through a simplified courtroom simulation with prosecutor, defense, and judge agents engaging in structured turn-taking. This provides iterative reasoning that extends beyond deterministic problems in math and coding, leveraging more accessible language and rhetorical structures \citep{castagna:24}. Each agent formulates private strategies before creating public utterances, incorporating elements of reflection, self-critique, and planning that align with recent work on argumentative LLMs \citep{sreedhar:24b}. We leverage established research on using LLMs as judges while acknowledging the limitations and biases inherent in such approaches \citep{schulhoff:24}.

\subsection{Explainability and Auditability in LaMAS}
A major challenge with using AI in high-stakes domains is their "black box" nature, which limits explainability and makes it difficult to verify alignment with human values. Multi-agent systems offer a potential solution by externalizing reasoning into observable agent interactions. AgenticSimLaw generates complete transcripts that record all utterances, private planning strategies, and justifications for final predictions, providing a highly interpretable form of eXplainable AI (XAI). Similar systems have been used to support medical reasoning for caregivers \citep{hong:24b}, demonstrating the practical value of logged agent interactions.

However, it is critical to distinguish between plausible and faithful explanations. Our research shows that self-reported prediction confidence does not correlate with performance accuracy, meaning these debate transcripts should be treated as plausible explanations of agent reasoning processes rather than objective evidence of correctness. This aligns with broader concerns in XAI literature about post-hoc rationalization. The value of our transcripts lies in enabling human auditors to trace the decision-making process, identify potential biases or errors in reasoning, and understand which features influenced the outcome—not in providing guarantees of logical soundness.

\section{Methodology}

\subsection{Dataset and Preprocessing}
Our tabular dataset consists of 1412 cases with 28 columns: one target label (True or False for 3-year rearrest) and 27 features. The feature set includes demographics, education, employment, family, drug use, religion, relationships, and criminal history collected between 1997-2002 from the National Longitudinal Survey of Youth (NLSY97). The target label was unbalanced, with 72\% "NO" and 28\% "YES" for recidivism. Feature distributions were also varied; for example, sex was nearly balanced (51\% male/49\% female), while racial/ethnic composition was as follows: 51.93\% non-black, non-Hispanic; 25.99\% black; 21.16\% Hispanic; and 0.92\% mixed non-Hispanic. Our task was binary classification based on a natural language narrative. This narrative was generated by concatenating all case facts into a string in the form of '<feature> is <value>' (e.g., 'sex is male'), which was then injected into the prompt. The dataset was split into a 60\% training, 20\% validation, and 20\% test set, with the test set used exclusively for evaluation.

\subsection{Experimental Setup}
All simulations were run in parallel on a local Ryzen 9 AMD PC with 128GB of RAM and dual NVIDIA 3090 GPUs (48GB VRAM total), utilizing Ollama version 0.5.7 with GPU acceleration via CUDA 12.6 and Python 3.10.12. Models were served using the ollama library version 0.4.5 with default hyperparameters. The temperature was set to 0.0 for single-turn prompting to ensure deterministic output and to 0.7 for the multi-turn MAD simulations to explore more creative and varied reasoning paths.

We grouped models into three ensembles of different sizes to ensure generalizability of our findings: a primary ensemble of 16 models with 7--14b parameters (Table~\ref{tab:small_model_ensemble_dual_col}), and two larger ensembles of 37 and 81 models with 0.5--72b parameters (Appendix~\ref{appendix:mad}). All models are 4-bit quantized (\texttt{q4\_K\_M}) unless otherwise noted and were sourced directly from ollama.ai \citep{ollama:25}. Models were selected for their popularity, performance, and recency on the ollama.ai and HuggingFace Open LLM Leaderboards \citep{huggingface:25a}. Our selection included uncensored models (e.g., Dolphin 3), older models for comparison (e.g., Llama 3.1), and models specifically focused on reasoning (e.g., DeepSeek-r1, Marco-o1).

\begin{table}[t]
  \centering
  \begin{tabular}{l c @{\hspace{2em}} l c}
    \hline
    \textbf{Model} & \textbf{Size} & \textbf{Model} & \textbf{Size} \\
    \hline
    aya-expanse   & 8b  & hermes 3    & 8b  \\
    deepseek-r1   & 7b  & llama 3.1   & 8b  \\
    dolphin 3     & 8b  & marco-o1    & 7b  \\
    exaone 3.5    & 8b  & mistral     & 7b  \\
    falcon 3      & 7b  & olmo 2      & 7b  \\
    gemma 2       & 9b  & phi 4       & 14b \\
    glm 2         & 9b  & qwen 2.5    & 7b  \\
    granite 3.1   & 8b  & tulu 3      & 8b  \\
    \hline
  \end{tabular}
  \caption{Small Model Ensemble}
  \label{tab:small_model_ensemble_dual_col}
\end{table}

\subsection{Reasoning Methodologies}
We benchmarked two distinct reasoning methods. A baseline (StandardLLM) consisted of three increasingly complex prompts designed to elicit more sophisticated reasoning. The prompt types were: (a) a minimal \textit{zero-shot} prompt requesting an immediate prediction, (b) a \textit{chain of thought (CoT)} prompt that elicits reasoning steps before a prediction, and (c) an \textit{n-shot CoT} prompt that injects n=30 labeled examples before requesting reasoning steps and a prediction. The prompt templates are shown in Appendix~\ref{appendix:standardllm_prompt}.
A second new role-structured, multi-agent evaluation framework (AgenticSimLaw), which uses a courtroom-style debate simulation to provide transparent and controllable reasoning over tabular predictions. The simulation models a simplified US bench criminal trial with prosecutor, defense, and judge agents. The framework is designed to be domain-agnostic—while we instantiate it for recidivism prediction, the same role-structured protocol applies to civil litigation, medical case review, credit assessment, or any deliberative decision-making context where adversarial argument testing and procedural transparency are valued.

\subsection{AgenticSimLaw: Multi-Agent Interaction Protocol}
\begin{figure}[t]
  \includegraphics[width=\columnwidth]{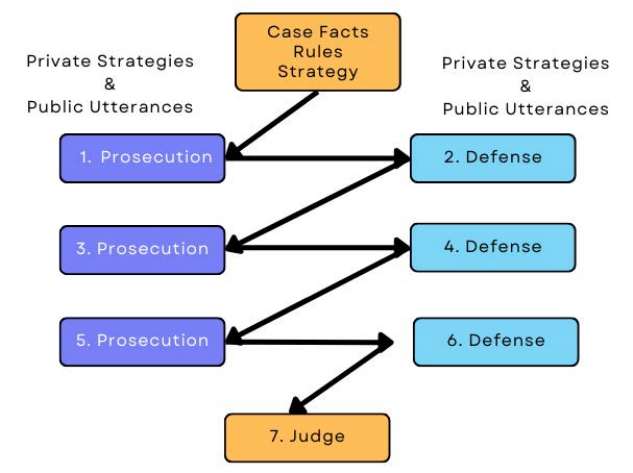}
  \caption{AgenticSimLaw Multi-Agent Interaction Framework. The diagram shows the role-structured 7-turn protocol with private strategy formulation (internal monologue) and public utterances (debate statements). All agent interactions and belief updates are logged for complete auditability.}
  \label{fig:agenticsimlaw_flowchart}
\end{figure}

The AgenticSimLaw framework implements a structured 7-turn interaction protocol that provides complete observability and control over the reasoning process. Figure~\ref{fig:agenticsimlaw_flowchart} shows the overall architecture. The protocol unfolds as follows:

\textbf{Prosecutor Opening (Turn 1)}: The prosecutor agent receives the case facts and privately formulates a strategy emphasizing high-risk factors before delivering a public opening statement arguing for recidivism.

\textbf{Defense Opening (Turn 2)}: The defense agent, having observed the prosecutor's opening, privately formulates a counter-strategy emphasizing protective factors before delivering a public opening statement arguing against recidivism.

\textbf{Judge Initial Belief Update}: The judge agent privately observes both opening statements and updates an internal belief state, explicitly tracking their current prediction (YES/NO), confidence level (0-100\%), and reasoning for that assessment.

\textbf{Prosecutor Rebuttal (Turn 3)}: The prosecutor privately strategizes a response to the defense opening, then delivers a public rebuttal that addresses defense arguments while reinforcing their position.

\textbf{Defense Rebuttal (Turn 4)}: The defense privately strategizes a counter-response, then delivers a public rebuttal addressing prosecutor arguments.

\textbf{Judge Mid-Debate Belief Update}: The judge privately updates their belief state after observing both rebuttals, again tracking prediction, confidence, and reasoning.

\textbf{Prosecutor Closing (Turn 5)}: The prosecutor delivers a final public closing argument, summarizing their case after privately strategizing the most compelling framing.

\textbf{Defense Closing (Turn 6)}: The defense delivers a final public closing argument, similarly prepared through private strategizing.

\textbf{Judge Final Verdict (Turn 7)}: The judge makes a final private belief update, then delivers a public verdict with explicit reasoning, prediction, and confidence level. This becomes the official simulation output.

Given the stochastic nature of LLMs, responses may be malformed or fail to complete. We implement a robust parsing strategy using both strict JSON methods and permissive regex patterns to capture valid predictions. If an agent fails to produce a parseable response after the primary attempt, we log the failure and either skip that turn (for intermediate utterances) or mark the simulation as incomplete (for the final judge verdict). All parsing failures and recovery attempts are logged for post-hoc analysis.

Every component of the interaction is logged: all public utterances, all private strategy formulations, all judge belief updates (with timestamps), and all API metadata (token counts, latency, temperature settings). This creates a complete audit trail that enables detailed analysis of agent behavior, belief evolution, and decision-making patterns. The logs serve dual purposes: (1) providing human-readable explanations for individual predictions, and (2) enabling systematic profiling of model behaviors across hundreds of simulations to identify consistent biases, failure modes, or reasoning patterns.

The AgenticSimLaw simulation, with its ~9,100 total tokens per run, represents a significant increase in test-time compute over single-shot CoT prompting (~500-800 tokens). This is a core aspect of our investigation into the benefits of structured multi-turn linguistic reasoning for LaMAS. The token investment buys three key properties: (1) \textbf{stability}—more consistent performance across metrics, (2) \textbf{transparency}—complete visibility into multi-step reasoning, and (3) \textbf{controllability}—explicit ability to modify agent roles, turn structure, or interaction rules. Future work could explore more efficient variants (e.g., reviewer-style aggregation, belief-based consensus) that reduce token costs while preserving key observability properties.

\subsection{Evaluation and Metrics}
For the StandardLLM experiments, each unique model and prompt type combination was tested on a pseudo-randomly selected subset of 150 cases from the test set. Due to the much higher compute costs, we ran 100 simulations for each of the 16 models in the small ensemble for AgenticSimLaw. The results of these LLM-based approaches are compared against a baseline of traditional statistical machine learning models (e.g., XGBoost) and specialized Tabular LLMs (e.g., TabPFN), which are detailed in the results section.

Given the high rate of malformed API responses, we used a robust parsing strategy that included both stricter JSON methods and more permissive regex patterns to capture as many valid predictions as possible. For each simulation, we logged all public utterances, the private evolution of internal belief states, and the judge's final opinion and reasoning. An API request to the Ollama server allowed us to parse the response and compare the prediction with the ground truth label. For each unique combination of model and prompt type, summary statistics were compiled, including mean prediction accuracy, F1 score, confusion matrix, and API metadata for calculating compute resources (e.g., execution time and token counts).

\subsection{Traditional Baseline Models}
To provide a comprehensive benchmark, we compared our LLM-based reasoning methodologies against both traditional statistical models and a specialized tabular LLM. The traditional models were evaluated using PyCaret, an open-source, low-code machine learning library for end-to-end model management. PyCaret was used to train and evaluate a wide range of classical GOFAI models, including Logistic Regression, Ridge Classifier, and Gradient Boosting Classifier, on our dataset using a cross-validation approach. The models were sorted by their Area Under the Curve (AUC) score, and their performance metrics (Accuracy, F1, Recall, etc.) were recorded to establish a strong performance baseline.

We also evaluated TabPFN, a Transformer-based model specifically designed for tabular data prediction in low-data regimes \citep{liu:25a}. The performance of TabPFN on our dataset was measured using its best hyperparameters, and its key metrics, including accuracy, AUC, and the confusion matrix, were recorded. These baseline results are presented in Section 4 and provide a clear reference point for evaluating the performance of our LLM-based approaches.

\section{Results}

\subsection{Performance of Traditional Baselines}
To benchmark our LLM approaches, we first evaluated standard tabular ML models produced by \textsc{PyCaret}'s compare/stack pipeline (Table~\ref{tab:pycaret_leaderboard}). On our recidivism dataset, the highest accuracy came from Random Forest (0.7606), while Extra Trees reached the best AUC (0.6559). Gradient Boosting obtained the top F1 (0.7190). Linear models (Ridge, Logistic Regression, LDA) were competitive on precision but trailed on AUC/accuracy; Naive Bayes and QDA underperformed substantially.

We also evaluated \texttt{TabPFN}, a specialized LLM for tabular data (Table~\ref{tab:tabpfn_baseline}). Despite reasonable AUC (0.6410) and accuracy (0.7011), \texttt{TabPFN} predicted no positives (recall \(=0\)), yielding \(\mathrm{F1}=0\) and a degenerate confusion matrix. This underscores that specialized tabular LLMs do not guarantee robust performance under our class imbalance and feature regime.

\begin{table*}[t]
\centering
\setlength{\tabcolsep}{3.5pt}\renewcommand{\arraystretch}{1.05}
\resizebox{\textwidth}{!}{%
\begin{tabular}{l l c c c c c c c c}
\hline
\textbf{ID} & \textbf{Model} & \textbf{Acc} & \textbf{AUC} & \textbf{Rec} & \textbf{Prec} & \textbf{F1} & \textbf{Kappa} & \textbf{MCC} & \textbf{TT (s)} \\
\hline
et       & Extra Trees Classifier        & 0.7551 & 0.6559 & 0.7551 & 0.7025 & 0.7107 & 0.1249 & 0.1438 & 0.4570 \\
catboost & CatBoost Classifier           & 0.7460 & 0.6449 & 0.7460 & 0.7008 & 0.7102 & 0.1295 & 0.1433 & 6.5810 \\
rf       & Random Forest Classifier      & \textbf{0.7606} & 0.6448 & \textbf{0.7606} & 0.6929 & 0.7020 & 0.0939 & 0.1142 & 0.5110 \\
gbc      & Gradient Boosting Classifier  & 0.7461 & 0.6375 & 0.7461 & 0.7135 & \textbf{0.7190} & 0.1695 & 0.1816 & 0.7500 \\
ridge    & Ridge Classifier              & 0.6187 & 0.6120 & 0.6187 & 0.7080 & 0.6442 & 0.1403 & 0.1580 & 0.3750 \\
lda      & Linear Discriminant Analysis  & 0.6131 & 0.6103 & 0.6131 & 0.7064 & 0.6393 & 0.1324 & 0.1512 & 0.2670 \\
lr       & Logistic Regression           & 0.6298 & 0.6088 & 0.6298 & 0.7098 & 0.6538 & 0.1524 & 0.1672 & 0.5370 \\
xgboost  & Extreme Gradient Boosting     & 0.7368 & 0.6015 & 0.7368 & 0.6919 & 0.7024 & 0.1091 & 0.1211 & 0.6750 \\
ada      & AdaBoost Classifier           & 0.7036 & 0.5970 & 0.7036 & 0.6791 & 0.6881 & 0.1016 & 0.0997 & 0.4550 \\
nb       & Naive Bayes                   & 0.2910 & 0.5854 & 0.2910 & 0.6425 & 0.2328 & -0.0029 & -0.0069 & 0.2680 \\
dt       & Decision Tree Classifier      & 0.6445 & 0.5299 & 0.6445 & 0.6638 & 0.6527 & 0.0554 & 0.0560 & 0.2700 \\
svm      & SVM (Linear Kernel)           & 0.3431 & 0.5231 & 0.3431 & 0.2908 & 0.2189 & -0.0013 & 0.0014 & 0.2820 \\
qda      & Quadratic Discriminant Analysis & 0.2763 & 0.5167 & 0.2763 & 0.6267 & 0.2030 & -0.0065 & -0.0190 & 0.2710 \\
knn      & K Neighbors Classifier        & 0.4789 & 0.4789 & 0.4789 & 0.6350 & 0.5153 & -0.0144 & -0.0186 & 0.4580 \\
\hline
\end{tabular}%
}
\caption{PyCaret baseline leaderboard on the recidivism task. Best per-metric: accuracy/recall (RF), AUC (ET), F1 (GBC).}
\label{tab:pycaret_leaderboard}
\end{table*}

\begin{table}[t]
\centering
\setlength{\tabcolsep}{5pt}\renewcommand{\arraystretch}{1.05}
\begin{tabular}{l c}
\hline
\textbf{Metric} & \textbf{Value} \\
\hline
AUC & 0.6410 \\
Accuracy & 0.7011 \\
Recall & 0.0000 \\
F1 Score & 0.0000 \\
Confusion Matrix & \texttt{[[319, 0]; [136, 0]]} \\
\hline
\end{tabular}
\caption{\texttt{TabPFN} baseline. Best hyperparameters: \texttt{learning\_rate=0.01}, \texttt{max\_depth=3}, \texttt{n\_estimators=100}.}
\label{tab:tabpfn_baseline}
\end{table}

\subsection{StandardLLM Performance Metrics}
All LLM models were benchmarked using accuracy and F1 score metrics. Due to the complexity of our prompts and the stochastic nature of LLMs, simulations often produced malformed responses or failed to complete. However, verbal fluency and reasoning were remarkably consistent, and performance was well above random, as seen in Table~\ref{tab:standardllm_performance}.

A central hypothesis of LLM reasoning is that performance should increase when shifting from zero-shot prompting to CoT and then to \(n\)-shot CoT, as more information is provided. However, for our ensemble of small (7--14b) models, a near reversal of this order occurred. Zero-shot prompting achieved the highest accuracy, while \(n\)-shot CoT was competitive on F1 scores. This suggests two non-exclusive explanations: (1) zero-shot prompts may have randomly overfit the data, and (2) smaller models may lack the capacity to fully exploit the greater information and complexity of more sophisticated prompts. Note that the uncensored Dolphin~3 and Gemma~2 models performed well, in contrast to the poor performance of typically higher-ranked models, suggesting that performance on this challenging task does not generalize well from standard benchmarks.

Table~\ref{tab:standardllm_performance} shows that \(n\)-shot CoT had a slight advantage over zero-shot prompts in terms of stability, as measured by the consistency of performance across different metrics. The slightly higher F1 performance suggests that small models can benefit from the 30 examples injected into the \(n\)-shot CoT prompt. An interesting finding is that the highest F1 performers are generally a mix of older models and those that excel on traditional benchmarks, while newer reasoning models like Marco-o1 perform in the middle to lower ranks. This reinforces the idea that, in the small model regime, simple zero-shot prompts can sometimes outperform models explicitly trained for reasoning on complex tasks, but such performance must be carefully validated against overfitting.

The top three models for accuracy were the uncensored Dolphin~3, Qwen~2.5, and Marco-o1. This may suggest that human alignment in these models is less focused on debiasing factors, which are a greater concern in Western cultures. For example, the relative ranking of Dolphin~3 compared to the uncensored Llama model it is based on (accuracy +11, F1 score -3) provides some evidence that anti-bias alignment may slightly decrease performance. The highly ranked DeepSeek-r1 model performed surprisingly poorly, perhaps because its bias for long verbal responses made it difficult to parse successfully and prone to producing malformed responses.

\begin{table}[t]
\centering
\setlength{\tabcolsep}{4pt}\renewcommand{\arraystretch}{1.05}
\resizebox{\columnwidth}{!}{%
\begin{tabular}{lcccccc}
\hline
\textbf{Model} & \multicolumn{2}{c}{\textbf{Zero-shot}} &
\multicolumn{2}{c}{\textbf{CoT}} &
\multicolumn{2}{c}{\(\mathbf{n}\)-\textbf{shot CoT}} \\
\hline
& Acc & F1 & Acc & F1 & Acc & F1 \\
\hline
dolphin 3 & 0.77 & 0.65 & 0.74 & 0.62 & 0.71 & 0.64 \\
qwen 2.5 & 0.74 & 0.60 & 0.67 & 0.58 & 0.70 & 0.62 \\
marco-o1 & 0.73 & 0.59 & 0.69 & 0.55 & 0.71 & 0.61 \\
llama 3.1 & 0.71 & 0.62 & 0.68 & 0.54 & 0.69 & 0.59 \\
gemma 2 & 0.70 & 0.53 & 0.66 & 0.52 & 0.72 & 0.58 \\
deepseek-r1 & 0.65 & 0.45 & 0.62 & 0.41 & 0.68 & 0.51 \\
\hline
\end{tabular}%
}
\caption{StandardLLM performance (small ensemble). Accuracies (Acc) and F1 scores across three prompting regimes.}
\label{tab:standardllm_performance}
\end{table}

\subsection{AgenticSimLaw Performance Metrics}
In contrast to our StandardLLM approach, traditional open LLM benchmark rankings do not reliably predict performance on our MAD courtroom reasoning task. Moreover, the AgenticSimLaw F1 scores were more consistent across models. While the greater test-time compute of MAD simulations resulted in negligible improvements over the best StandardLLM zero-shot performance metrics, structured multi-agent linguistic debate offers notable improvements in \textbf{stability} and \textbf{generalizability} compared to CoT and \(n\)-shot CoT prompting. We define stability as more consistent performance across different evaluation metrics (accuracy and F1 score), suggesting a more robust and less brittle reasoning process.

Table~\ref{tab:agenticsimlaw_performance} shows that the rank order for accuracy vs.\ F1 score for AgenticSimLaw models is somewhat reversed from the StandardLLM order. Newer, highly ranked reasoning models like Qwen~2.5, Marco-o1, and Phi-4 are top in accuracy, but their F1 scores are middling. Conversely, older, lower-ranked models like Llama~3.1, Mistral, and Hermes~3 show higher F1 scores. This suggests that these simpler models can benefit most from the structured linguistic debate, which forces a more comprehensive exploration of the case facts.

There is little correlation between the performance rankings under StandardLLM and AgenticSimLaw methodologies. The clustering of older, less performant models at the top of AgenticSimLaw F1 scores suggests that these simple models can benefit most from the explicit multi-round, multi-agent structured linguistic debate simulation. These results highlight that traditional LLM leaderboard rankings do not generalize well to our reasoning tasks using either traditional CoT prompting or MAD simulations.

\begin{table}[t]
\centering
\setlength{\tabcolsep}{10pt}\renewcommand{\arraystretch}{1.1}
\resizebox{\columnwidth}{!}{%
\begin{tabular}{lcc}
\hline
\textbf{Model} & \textbf{Accuracy} & \textbf{F1 Score} \\
\hline
qwen 2.5 & 0.76 & 0.60 \\
marco-o1 & 0.75 & 0.61 \\
dolphin 3 & 0.75 & 0.58 \\
phi 4 & 0.73 & 0.56 \\
llama 3.1 & 0.71 & 0.62 \\
gemma 2 & 0.70 & 0.64 \\
deepseek-r1 & 0.68 & 0.52 \\
\hline
\end{tabular}%
}
\caption{AgenticSimLaw performance (small ensemble).}
\label{tab:agenticsimlaw_performance}
\end{table}

\subsection{Commercial LLM Performance}
We also tested four SOTA commercial LLMs to validate the assumption that more complex and informative reasoning prompts should result in better reasoning and predictions. Table~\ref{tab:sota_commercial} shows these models all trend in this direction, although only OpenAI's \texttt{o3-mini} model demonstrated this with realistic statistical distributions. Haiku-3.5's higher metrics in some cases are discounted by the fact that it only predicted \texttt{NO} for every API call in its CoT and \(n\)-shot CoT runs, resulting in an F1 score of 0.0 despite its high accuracy. This further validates the importance of using both accuracy and F1 score for a comprehensive evaluation, especially in imbalanced datasets.

\begin{table}[t]
\centering
\setlength{\tabcolsep}{5pt}\renewcommand{\arraystretch}{1.05}
\resizebox{\columnwidth}{!}{%
\begin{tabular}{l l c c}
\hline
\textbf{Model} & \textbf{Prompt} & \textbf{Accuracy} & \textbf{F1} \\
\hline
\texttt{gpt-4o-mini} & zero-shot & 0.48 & 0.60 \\
\texttt{gpt-4o-mini} & cot & 0.71 & 0.50 \\
\texttt{gpt-4o-mini} & n-shot & 0.47 & 0.49 \\
\hline
\texttt{o3-mini} & zero-shot & 0.49 & 0.63 \\
\texttt{o3-mini} & cot & 0.53 & 0.60 \\
\texttt{o3-mini} & n-shot & 0.70 & 0.70 \\
\hline
\texttt{haiku-3.5} & zero-shot & 0.34 & 0.21 \\
\texttt{haiku-3.5} & cot & 0.71 & 0.00 \\
\texttt{haiku-3.5} & n-shot & 0.71 & 0.00 \\
\hline
\texttt{sonnet-3.5} & zero-shot & 0.43 & 0.44 \\
\texttt{sonnet-3.5} & cot & 0.53 & 0.68 \\
\texttt{sonnet-3.5} & n-shot & 0.57 & 0.72 \\
\hline
\end{tabular}%
}
\caption{SOTA commercial LLM performance across prompting regimes.}
\label{tab:sota_commercial}
\end{table}

\subsection{Compute Cost vs. Transparency Tradeoff}
A critical consideration for deploying multi-agent systems is the computational overhead relative to single-agent approaches. AgenticSimLaw's 7-turn protocol requires approximately \textbf{11--14$\times$ more tokens} than single-turn CoT prompting (9,100 vs. 650--800 tokens per prediction). On our hardware, this translates to roughly \textbf{8--12$\times$ longer wall-clock time} per case. However, this cost delivers three key benefits that justify the overhead for high-stakes applications:

\textbf{(1) Metric Stability}: AgenticSimLaw exhibits lower variance in F1 scores across models (std.~dev.~0.04 vs.~0.08 for StandardLLM zero-shot), indicating more predictable performance that is less sensitive to specific model quirks or random initialization.

\textbf{(2) Transparency}: The complete debate transcript provides human auditors with a step-by-step reasoning trace, whereas single-turn CoT outputs offer only a condensed justification that may omit critical decision factors.

\textbf{(3) Control}: The modular 7-turn structure allows researchers to modify agent roles, adjust argumentation strategies, or inject domain-specific rules without retraining models—providing a level of test-time configurability that internal reasoning models like o3-mini cannot match.

For applications where these properties are not required (e.g., low-stakes classification with well-calibrated models), simpler prompting strategies remain more efficient. However, for domains demanding auditability, procedural fairness, and human oversight—such as criminal justice, medical diagnosis, or financial lending—the compute-transparency tradeoff strongly favors structured multi-agent orchestration. Future work could explore efficiency optimizations inspired by reviewer-style MAS, such as reducing inter-agent communication rounds or using belief-based aggregation while preserving the core observability benefits.

\section{Discussion and Responsible Use}

Standard benchmarks have long been criticized as poor predictors of real-world performance. This paper further explores this by using small 7--14b open LLMs on the extremely challenging task of young adult recidivism prediction. As shown by our baselines, statistical ML models like XGBoost typically far outperform general LLM linguistic reasoning on tabular data. Our results show that LLM performance on this task is highly specific to the particular combination of data, model type, size, and reasoning methodology.

Within our resource constraints, we found that neither top models on traditional benchmark leaderboards nor the recent crop of fine-tuned reasoning models have consistent advantages. In fact, some of the oldest and lowest-performing models benefited most from MAD simulations when ranked by F1 score, suggesting the structured debate helps compensate for their simpler architectures. Conversely, the newest and highly regarded DeepSeek-r1 7b performed near the bottom, possibly due to a bias towards long verbal responses that hinders parsing. Simple zero-shot prompting exhibited the highest accuracy, but its significant performance variance suggests this may be due to overfitting and that the data, task/prompt, model type, and size must be jointly optimized to generalize. Finally, we demonstrate with four SOTA commercial LLMs that much more powerful models do, in fact, benefit from more complex reasoning prompts, highlighting the current limitations of smaller open-source LLMs.

\subsection{Responsible Deployment Considerations}
\textbf{This work is intended for research, benchmarking, and transparency analysis—not for operational deployment in sentencing, parole, or any consequential criminal justice decisions.} Recidivism prediction in real-world systems exhibits several properties that our experimental setup cannot capture:

Algorithmic predictions can alter outcomes through feedback loops (e.g., increased surveillance of high-risk individuals may cause arrests unrelated to criminal propensity) \citep{angwin:16}. Recidivism patterns change over time due to policy shifts, economic conditions, and social movements. Our static NLSY97 dataset (1997-2002) would not reflect contemporary patterns.

While our dataset includes demographic features, we have not conducted rigorous disparate impact analysis across race, gender, or socioeconomic strata. Such analysis is mandatory before any deployment and must meet legal standards of non-discrimination. Even if performance were adequate, algorithmic recommendations should serve only as one input to human decision-makers who consider contextual factors, legal standards, and individual circumstances that tabular data cannot capture.

AgenticSimLaw's value lies in providing a transparent scaffold for auditing these concerns. The complete debate transcripts allow domain experts to trace exactly how demographic features, criminal history, and social factors influenced the final prediction—enabling bias detection and accountability that black-box systems cannot provide. However, transparency does not equal fairness or correctness. Our framework should be viewed as a \textbf{research tool for studying multi-agent reasoning} rather than a production system for high-stakes decisions.

While this is an exploratory introduction to a novel approach for applying MAD simulations to tabular LLM predictions, our findings are preliminary. The complete logs from our experiments, including reasoning transcripts, provide valuable insight into the explainability of the MAD approach. For instance, the judge's internal monologues often reveal the trade-offs being considered, such as balancing a strong family background against a history of drug use, providing a level of procedural transparency that is absent in traditional statistical models. This transparency is a key advantage of the MAD methodology, offering a human-readable trace of the decision-making process. Future work, with greater resources, could multiply these studies to define confidence ranges and statistically validate these findings.

\section{Conclusion}

We have introduced AgenticSimLaw, a role-structured multi-agent evaluation framework that provides transparent, controllable, and auditable reasoning for high-stakes tabular decision-making. Through a courtroom-style debate protocol with explicit agent roles, a 7-turn interaction sequence, and complete logging of utterances and belief updates, our framework addresses core challenges in LLM-based multi-agent systems: organization through structured roles, observability through logged interactions, and responsibility through explicit deployment constraints.

Our experiments across almost 90 unique model and prompt combinations demonstrate that structured multi-agent debate improves reasoning stability and generalizability compared to single-agent Chain-of-Thought prompting, evidenced by stronger correlation between accuracy and F1 metrics. Beyond performance gains, AgenticSimLaw offers fine-grained control over test-time reasoning, generates human-readable debate transcripts as plausible explanations of agent behavior, and enables systematic profiling of model reasoning patterns. The framework is domain-agnostic—while we instantiate it for recidivism prediction to stress-test reasoning under ethical complexity, the same protocol applies to medical case review, credit assessment, policy analysis, or any deliberative context valuing procedural transparency.

Using smaller open LLMs with AgenticSimLaw also offers practical benefits: a highly explainable, local decision-support system that is private, customizable, reliable, fast, portable, auditable, and inexpensive. It allows researchers to make informed choices on the tradeoff between alignment and performance and provides feedback to improve alignment training. The compute-transparency tradeoff analysis shows that the framework's 11--14$\times$ token overhead compared to single-turn prompting delivers critical observability and control properties that black-box approaches cannot provide—a justified cost for high-stakes applications requiring human oversight.

We emphasize that AgenticSimLaw is designed as a research and evaluation tool, not for operational deployment in consequential decision-making without rigorous fairness analysis, legal review, and human supervision. The framework's transparent interaction logs make it particularly suitable for studying bias, evaluating alignment, and developing best practices for responsible multi-agent system design—contributing to the broader goal of trustworthy, verifiable, and human-aligned LaMAS.

\section{Ethics Statement}

Given the high-stakes implications of recidivism prediction within the judicial system, we recognize the profound ethical responsibilities associated with this work. The NLSY97 dataset is publicly available, and no personally identifiable information was used. We acknowledge that algorithmic predictions, particularly in sensitive domains like criminal justice, can perpetuate or amplify existing societal biases. 

\textbf{Non-Deployment Declaration}: This research is intended exclusively for academic study, framework development, and transparency analysis. \textbf{AgenticSimLaw should not be deployed for operational sentencing, parole, probation, or any consequential criminal justice decisions.}

Recognizing the risk of "anti-bias" overtuning, where attempts to correct biases can inadvertently introduce new ones or reduce predictive utility, we suggest that some of these biases may explain our findings regarding model performance differences. Our results should be interpreted as part of an ongoing dialogue on the responsible, ethical, and legal use of AI in high-stakes decision-making. We stress that debate transcripts, while providing transparency, should be treated as plausible rather than faithful explanations—they illuminate agent reasoning processes but do not guarantee logical soundness or eliminate bias.

The value of AgenticSimLaw lies in making reasoning observable and thus auditable, enabling researchers and practitioners to systematically identify failure modes, test for disparate impact, and refine agent behaviors. However, observability is a necessary but insufficient condition for fairness. We caution against over-reliance on any algorithmic system and emphasize the continued centrality of human judgment, contextual understanding, and legal accountability in high-stakes domains.

\section{Limitations and Future Work}

Anthropomorphizing terms like 'strategize', 'debate', and 'internal belief states' are descriptive shorthand for the functional aspects of our agents and simulation. They are not to be taken literally or to suggest that the authors are ascribing full and genuine human attributes to AI.

The primary focus of this paper is to develop a multi-agent evaluation framework that surfaces the reasoning limits of open LLMs, particularly in the 7-14B parameter range popular with local deployments, when applied to challenging tabular prediction tasks. While our extensive ensemble provides a broad overview, a key limitation is the reliance on a single binary classification task. Future work should expand by surveying diverse tabular datasets and introducing more complex multi-label or regression tasks. Additionally, future research should perform more statistically rigorous experiments, focusing on fewer key model and prompt combinations with repeated runs to develop statistical confidence metrics and further validate our findings.

Given the stochastic nature of LLMs, their inherent innumeracy, and 1D linear attention mechanisms, tabular predictions currently excel only in low-data regimes, beyond which statistical ML models like XGBoost remain SOTA. To truly exploit LLM strengths would involve incorporating multi-step processing, synthesizing multi-modal data, and employing agentic tool use with complementary abilities (e.g., a Python interpreter with AutoML to drive XGBoost models). Future work could also explore efficiency optimizations inspired by reviewer-style or belief-based multi-agent systems, potentially reducing the 7-turn protocol to fewer interactions while preserving key observability properties. Finally, further analysis is needed to understand counterintuitive performance results, such as the unexpected underperformance of certain models like DeepSeek-r1, and whether this reflects genuine reasoning limitations or merely artifacts of structured output formatting requirements.


\bibliography{aaai2026}

@article{angwin:16,
  author  = "Angwin, Julia and Larson, Jeff and Mattu, Surya and Kirchner, Lauren",
  title   = "{Machine Bias: There's Software Used Across the Country to Predict Future Criminals. And It's Biased Against Blacks}",
  journal = "ProPublica",
  year    = 2016,
}

@misc{castagna:24,
  title   = "{Critical-Questions-of-Thought: Steering {LLM} reasoning with Argumentative Querying}",
  author  = "Castagna, Federico and Rahwan, Iyad and Parsons, Simon",
  year    = 2024,
  eprint  = "2412.15177",
  archivePrefix = "arXiv",
  primaryClass = "cs.AI",
}

@misc{cottier:24b,
  author  = "Cottier, Ben",
  title   = "{How Much Does It Cost to Train Frontier AI Models?}",
  year    = 2024,
  howpublished = "\url{https://epoch.ai/blog/how-much-does-it-cost-to-train-frontier-ai-models}",
  publisher = "Epoch AI",
  note    = "Accessed [Insert Date]",
}

@misc{deepseek:25,
  title   = "{DeepSeek-R1: Incentivizing Reasoning Capability in LLMs via Reinforcement Learning}",
  author  = "{DeepSeek-AI} and Guo, Daya and Yang, Dejian and Zhang, Haowei and Song, Junxiao and Zhang, Ruoyu and Xu, Runxin and Zhu, Qihao and Ma, Shirong and Wang, Peiyi and Bi, Xiao and Zhang, Xiaokang and Yu, Xingkai and Wu, Yu and Wu, Z. F. and Gou, Zhibin and Shao, Zhihong and Li, Zhuoshu and Gao, Ziyi and Liu, Aixin and Xue, Bing and Wang, Bingxuan and Wu, Bochao and Feng, Bei and Lu, Chengda and Zhao, Chenggang and Deng, Chengqi and Zhang, Chenyu and Ruan, Chong and Dai, Damai and Chen, Deli and Ji, Dongjie and Li, Erhang and Lin, Fangyun and Dai, Fucong and Luo, Fuli and Hao, Guangbo and Chen, Guanting and Li, Guowei and Zhang, H. and Bao, Han and Xu, Hanwei and Wang, Haocheng and Ding, Honghui and Xin, Huajian and Gao, Huazuo and Qu, Hui and Li, Hui and Guo, Jianzhong and Li, Jiashi and Qiu, Junjie and Li, Junlong and Cai, J. L. and Ni, Jiaqi and Liang, Jian and Chen, Jin and Dong, Kai and Hu, Kai and Gao, Kaige and Guan, Kang and Huang, Kexin and Yu, Kuai and Wang, Lean and Zhang, Lecong and Zhao, Liang and Wang, Litong and Zhang, Liyue and Xu, Lei and Xia, Leyi and Zhang, Mingchuan and Zhang, Minghua and Tang, Minghui and Li, Meng and Wang, Miaojun and Li, Mingming and Tian, Ning and Huang, Panpan and Zhang, Peng and Wang, Qiancheng and Chen, Qinyu and Du, Qiushi and Ge, Ruiqi and Pan, Ruizhe and Ji, Run and Chen, R. J. and Jin, R. L. and Chen, Ruyi and Lu, Shanghao and Zhou, Shangyan and Chen, Shanhuang and Ye, Shengfeng and Wang, Shiyu and Yu, Shuiping and Zhou, Shunfeng and Pan, Shuting and Li, S. S. and Zhou, Shuang and Wu, Shaoqing and Ye, Shengfeng and Yun, Tao and Pei, Tian and Sun, Tianyu and Wang, T. and Zeng, Wangding and Zhao, Wanjia and Liu, Wen and Liang, Wenfeng and Gao, Wenjun and Yu, Wenqin and Zhang, Wentao and Xiao, W. L. and An, Wei and Liu, Xiaodong and Wang, Xiaohan and Chen, Xiaokang and Nie, Xiaotao and Cheng, Xin and Liu, Xin and Xie, Xin and Liu, Xingchao and Yang, Xinyu and Liu, Xinyuan and Su, Xuecheng and Lin, Xuheng and Li, X. Q. and Jin, Xiangyue and Shen, Xiaojin and Chen, Xiaosha and Sun, Xiaowen and Wang, Xiaoxiang and Song, Xinnan and Zhou, Xinyi and Wang, Xianzu and Shan, Xinxia and Li, Y. K. and Wang, Y. Q. and Wei, Y. X. and Zhang, Yang and Xu, Yanhong and Li, Yao and Zhao, Yao and Sun, Yaofeng and Li, Yaohui and Yu, Yi and Zhang, Yichao and Shi, Yifan and Xiong, Yiliang and He, Ying and Piao, Yishi and Wang, Yisong and Tan, Yixuan and Ma, Yiyang and Liu, Yiyuan and Guo, Yongqiang and Ou, Yuan and Wang, Yuduan and Gong, Yue and Zou, Yuheng and He, Yujia and Xiong, Yunfan and Luo, Yuxiang and You, Yuxiang and Liu, Yuxuan and Zhou, Yuyang and Zhu, Y. X. and Xu, Yanhong and Huang, Yanping and Li, Yaohui and Zheng, Yi and Zhu, Yuchen and Ma, Yunxian and Tang, Ying and Zha, Yukun and Yan, Yuting and Ren, Z. Z. and Ren, Zehui and Sha, Zhangli and Fu, Zhe and Xu, Zhean and Xie, Zhenda and Zhang, Zhengyan and Hao, Zhewen and Ma, Zhicheng and Yan, Zhigang and Wu, Zhiyu and Gu, Zihui and Zhu, Zijia and Liu, Zijun and Li, Zilin and Xie, Ziwei and Song, Ziyang and Pan, Zizheng and Huang, Zhen and Xu, Zhipeng and Zhang, Zhongyu and Zhang, Zhen",
  year    = 2025,
  eprint  = "2501.12948",
  archivePrefix = "arXiv",
  primaryClass = "cs.CL",
}

@article{dressel:18,
  author  = "Dressel, Julia and Farid, Hany",
  title   = "{The accuracy, fairness, and limits of predicting recidivism}",
  journal = "Science Advances",
  volume  = 4,
  number  = 1,
  pages   = "eaao5580",
  year    = 2018,
}

@misc{fang:24a,
  title   = "{Large Language Models(LLMs) on Tabular Data: Prediction, Generation, and Understanding - A Survey}",
  author  = "Fang, Xingbo and Xu, Wensheng and Tan, Fong Ching Alvin and Zhang, Jingbo and Hu, Zhen and Qi, Yifeng and Nickleach, Stephen and Socolinsky, Diego and Sengamedu, Srinivasan H. and Faloutsos, Christos",
  year    = 2024,
  eprint  = "2402.17944",
  archivePrefix = "arXiv",
  primaryClass = "cs.LG",
}

@misc{freedman:24b,
  title   = "{Argumentative Large Language Models for Explainable and Contestable Decision-Making}",
  author  = "Freedman, Gabriel and Dejl, Adam and Gorur, Dilara and Yin, Xiaoyu and Rago, Antonio and Toni, Francesca",
  year    = 2024,
  eprint  = "2405.02079",
  archivePrefix = "arXiv",
  primaryClass = "cs.AI",
}

@article{hollmann:25,
  title   = "{Accurate predictions on small data with a tabular foundation model}",
  author  = "Hollmann, Noah and Müller, Samuel G. and Purucker, Lennart and Krishnakumar, Arjun and Körfer, Max and Hoo, Shi Bin and Schirrmeister, Robin Tibor and Hutter, Frank",
  journal = "Nature",
  year    = 2025,
  volume  = 637,
  pages   = "319 - 326",
}

@misc{hong:24b,
  title   = "{ArgMed-Agents: Explainable Clinical Decision Reasoning with Large Language Models via Argumentation Schemes}",
  author  = "Hong, Seunghoon and Xiao, Liang and Zhang, Xiaodan and Chen, Jing",
  year    = 2024,
  eprint  = "2403.06294",
  archivePrefix = "arXiv",
  primaryClass = "cs.CL",
}

@misc{hu:21,
  title   = "{LoRA: Low-Rank Adaptation of Large Language Models}",
  author  = "Hu, Edward J. and Shen, Yelong and Wallis, Phillip and Allen-Zhu, Zeyuan and Li, Yuanzhi and Wang, Shean and Wang, Lu and Chen, Weizhu",
  year    = 2021,
  eprint  = "2106.09685",
  archivePrefix = "arXiv",
  primaryClass = "cs.CL",
}

@misc{huang:24c,
  title   = "{OlympicArena: Benchmarking Multi-discipline Cognitive Reasoning for Superintelligent AI}",
  author  = "Huang, Zhen and Wang, Zengzhi and Xia, Shijie and Li, Xuefeng and Zou, Haoyang and Xu, Ruijie and Fan, Run-Ze and Ye, Lyumanshan and Chern, Ethan and Ye, Yixin and Zhang, Yikai and Yang, Yuqing and Wu, Ting and Wang, Binjie and Sun, Shichao and Xiao, Yang and Li, Yiyuan and Zhou, Fan and Chern, Steffi and Qin, Yiwei and Ma, Yan and Su, Jiadi and Liu, Yixiu and Zheng, Yuxiang and Zhang, Shaoting and Lin, Dahua and Qiao, Yu and Liu, Pengfei",
  year    = 2024,
  eprint  = "2406.12753",
  archivePrefix = "arXiv",
  primaryClass = "cs.CL",
}

@misc{huggingface:25a,
  title   = "{Open LLM Leaderboard}",
  author  = "{Huggingface}",
  year    = 2025,
  howpublished = "\url{https://huggingface.co/spaces/open-llm-leaderboard/open_llm_leaderboard}",
  note    = "Accessed: 2025-02-08",
}

@misc{ibm:24,
  title   = "{IBM Granite 3.1 Language Models}",
  author  = "{IBM Research}",
  year    = 2024,
  howpublished = "\url{https://github.com/ibm-granite/granite-3.1-language-models}",
}

@misc{kaplan:20,
  title   = "{Scaling Laws for Neural Language Models}",
  author  = "Kaplan, Jared and McCandlish, Sam and Henighan, Tom and Brown, Tom B. and Chess, Benjamin and Child, Rewon and Gray, Scott and Radford, Alec and Wu, Jeffrey and Amodei, Dario",
  year    = 2020,
  eprint  = "2001.08361",
  archivePrefix = "arXiv",
  primaryClass = "cs.LG",
}

@misc{lambert:25,
  title   = "{Tulu 3: Pushing Frontiers in Open Language Model Post-Training}",
  author  = "Lambert, Nathan and Morrison, Jacob and Pyatkin, Valentina and Huang, Shengyi and Ivison, Hamish and Brahman, Faeze and Miranda, Lester James V. and Liu, Alisa and Dziri, Nouha and Lyu, Shane and Gu, Yuling and Malik, Saumya and Graf, Victoria and Hwang, Jena D. and Yang, Jiangjiang and Le Bras, Ronan and Tafjord, Oyvind and Wilhelm, Chris and Soldaini, Luca and Smith, Noah A. and Wang, Yizhong and Dasigi, Pradeep and Hajishirzi, Hannaneh",
  year    = 2025,
  eprint  = "2411.15124",
  archivePrefix = "arXiv",
  primaryClass = "cs.CL",
}

@misc{liu:25a,
  author  = "Liu, Siyang and others",
  title   = "{{TabPFN} Unleashed: A Scalable and Effective Solution to Tabular Classification Problems}",
  year    = 2025,
  eprint  = "2502.02527",
  archivePrefix = "arXiv",
  primaryClass = "cs.LG",
}

@misc{ma:24,
  title   = "{In-Context Data Distillation with TabPFN}",
  author  = "Ma, Junwei and Thomas, Valentin and Yu, Guangwei and Caterini, Anthony",
  year    = 2024,
  eprint  = "2402.06971",
  archivePrefix = "arXiv",
  primaryClass = "cs.LG",
}

@misc{ollama:25,
  title   = "{Model library}",
  author  = "{Ollama}",
  year    = 2025,
  howpublished = "\url{https://ollama.com/search}",
  note    = "Accessed: 2025-02-08",
}

@misc{teamolmo:25,
  title   = "{2 OLMo 2 Furious}",
  author  = "{Team OLMo} and Walsh, Pete and Soldaini, Luca and Groeneveld, Dirk and Lo, Kyle and Arora, Shane and Bhagia, Akshita and Gu, Yuling and Huang, Shengyi and Jordan, Matt and Lambert, Nathan and Schwenk, Dustin and Tafjord, Oyvind and Anderson, Taira and Atkinson, David and Brahman, Faeze and Clark, Christopher and Dasigi, Pradeep and Dziri, Nouha and Guerquin, Michal and Ivison, Hamish and Koh, Pang Wei and Liu, Jiacheng and Malik, Saumya and Merrill, William and Miranda, Lester James V. and Morrison, Jacob and Murray, Tyler and Nam, Crystal and Pyatkin, Valentina and Rangapur, Aman and Schmitz, Michael and Skjonsberg, Sam and Wadden, David and Wilhelm, Christopher and Wilson, Michael and Zettlemoyer, Luke and Farhadi, Ali and Smith, Noah A. and Hajishirzi, Hannaneh",
  year    = 2025,
  eprint  = "2501.00656",
  archivePrefix = "arXiv",
  primaryClass = "cs.CL",
}

@misc{openai:24,
  title   = "{Learning to reason with LLMs}",
  author  = "{OpenAI}",
  year    = 2024,
  howpublished = "\url{https://openai.com/index/learning-to-reason-with-llms/}",
  publisher = "OpenAI",
}

@misc{ruan:24,
  title   = "{Language Modeling on Tabular Data: A Survey of Foundations, Techniques and Evolution}",
  author  = "Ruan, Yucheng and Lan, Xiang and Ma, Jingying and Dong, Yizhi and He, Kai and Feng, Mengling",
  year    = 2024,
  eprint  = "2408.10548",
  archivePrefix = "arXiv",
  primaryClass = "cs.CL",
}

@misc{schulhoff:24,
  title   = "{The Prompt Report: A Systematic Survey of Prompting Techniques}",
  author  = "Schulhoff, Sophia and Ilie, Mihnea and Balepur, Nikhil and Kahadze, Koba and Liu, Andrew and Si, Chenglei and Li, Yiyang and Gupta, Aniket and Han, Hanwen and Schulhoff, Sophia and Dulepet, Pranav Sai and Vidyadhara, Sai and Ki, Donghyun and Agrawal, Shivam and Pham, Chau and Kroiz, Gabriel C and Li, Fangzhou and Tao, Huanqi and Srivastava, Ananya and Costa, Hugo D and Gupta, Shivam and Rogers, Megan L and Goncearenco, Igor and Sarli, Gabriele and Galynker, Igor and Peskoff, Daniel and Carpuat, Marine and White, Jennifer and Anadkat, Shivam and Hoyle, Alyssa M and Resnik, Philip",
  year    = 2024,
  eprint  = "2406.06608",
  archivePrefix = "arXiv",
  primaryClass = "cs.CL",
}

@unpublished{sreedhar:24b,
  title   = "{Simulating Human Strategic Behavior: Comparing Single and Multi-agent LLMs}",
  author  = "Sreedhar, Karthik and Chilton, Lydia B.",
  year    = 2024,
  note    = "Preprint",
}

@article{wang:23,
  title   = "{Learning to break: Knowledge-enhanced reasoning in multi-agent debate system}",
  author  = "Wang, Haozhe and Du, Xiaohan and Yu, Wenhao and Chen, Qian and Zhu, Kaizhu and Chu, Zhongjin and Yan, Liang and Guan, Yilong",
  journal = "Neurocomputing",
  volume  = 618,
  pages   = "129063",
  year    = 2023,
  publisher = "Elsevier",
}

@article{williams:24,
  author  = "Williams, Kristian and Papp, John and Brayne, Sarah",
  title   = "{Risk Assessment Tools and Racial Disparities: Evidence from COMPAS in Practice}",
  journal = "Social Problems",
  volume  = 71,
  number  = 2,
  pages   = "445--467",
  year    = 2024,
}

@article{pnas:24,
  author  = "Tjuatja, Lindia and Chen, Valerie and Wu, Tongshuang and Talwalkar, Ameet and Neubig, Graham",
  title   = "{LLMs Exhibit Human-Like Cognitive Biases at Systematic and Amplified Levels}",
  journal = "Proceedings of the National Academy of Sciences",
  volume  = 121,
  number  = 35,
  year    = 2024,
}

@misc{openai:24b,
  title   = "{Learning to Reason with LLMs}",
  author  = "{OpenAI}",
  year    = 2024,
  howpublished = "\url{https://openai.com/index/learning-to-reason-with-llms/}",
}

@misc{universalprm:24,
  title   = "{Rewarding Progress: Scaling Automated Process Verifiers for LLM Reasoning}",
  author  = "Setlur, Amrith and Nagpal, Chirag and Fisch, Adam and Geng, Xinyang and Eisenstein, Jacob and Agarwal, Rishabh and Agarwal, Alekh and Berant, Jonathan and Kumar, Aviral",
  year    = 2024,
  eprint  = "2410.08146",
  archivePrefix = "arXiv",
  primaryClass = "cs.LG",
}

\onecolumn

\appendix
\section{Appendix A: StandardLLM Prompt Templates and Data Sample}
\label{appendix:standardllm_prompt}

\subsection{Dataset Sample Represented as a String}
The person is an 18-year-old male who identifies himself as Hispanic. He has been arrested 6 times and has been convicted. He is 5 feet 0 inches and weighs between 150 and 174 lbs. He has never been married and is currently not living alone. He does not have a high school diploma or GED certificate and is not enrolled in college. He has held no jobs during the year. He has used drugs in the past four years (marijuana) and has not felt depressed over the past month. At a young age, he lived with his biological mother in a household of size 6, in an urban area. He considers that God has something to do with what happens to him personally.

\subsection{Prompt Templates}
\begin{verbatim}
prompts:
  prompt_persona: |
    You are a highly experienced criminal justice risk assessment expert
    employing advanced statistical and judicial knowledge.

  system1: |
    You are a risk assessment expert. Analyze the following risk factors 
    to determine if they indicate high risk.
    You must respond ONLY with a JSON object in exactly this format, 
    with no additional text:
    
    {
      "prediction": "YES" or "NO",
      "confidence": number between 0 and 100
    }
    Risk Factors to analyze:
    {risk_factors}

  cot: |
    You are a criminal risk assessment expert. Step back, take a deep breath 
    and carefully think step by step to assign a relative weight [low, medium, high] 
    to each risk factor before using them to collectively predict 3-year 
    rearrest recidivism:
    {risk_factors}
    After your analysis, you must respond ONLY with a JSON object in exactly 
    this format, with no additional text:
    
    {
      "risk_factors": [
        ("factor1", "low"|"medium"|"high", "textual reasoning"),
        ("factor2", "low"|"medium"|"high", "textual reasoning"),
        ...
      ],
      "prediction": "YES" or "NO", 
      "confidence": number between 0 and 100
    }

  cot-nshot: |
    You are a criminal risk assessment expert. First,analyze the following 
    example cases of risk factors and resulting 3 year rearrest recidivism outcome:
    {nshot_example_str}
    
    Next, step back and take a deep breath and carefully think step by step 
    to assign a relative weight [low, medium, high] to each risk factor before 
    using them to collectively predict 3-year rearrest recidivism for this case:
    {risk_factors}
    Based on the above n-shot examples and your weighted risk factor analysis 
    for this particular case, predict the 3-year rearrest recidivism outcome 
    for this case.
    Respond ONLY with a JSON object in exactly this format, with no additional text:
    
    {
      "risk_factors": [
        ("factor1", "low"|"medium"|"high", "textual reasoning"),
        ("factor2", "low"|"medium"|"high", "textual reasoning"),
        ...
      ],
      "prediction": "YES" or "NO",
      "confidence": number between 0 and 100
    }
\end{verbatim}

\clearpage

\appendix
\section{Appendix B: Sample Courtroom Debate Simulation}
\label{appendix:agenticsimlaw_transcript}

We present a sample debate simulation from Granite3.1-dense:8b (transcript\_row-123\_ver-3.json \citep{ibm:24}).

\subsection{Case Facts}
Age: 22, Prior Arrests: 2, Actual Recidivism: No

Additional characteristics: highest degree is not enrolled with high school degree; parent/guardian relationship at age 12 is Unknown; married/cohabitation status is never married, not cohabiting; resident locale is Unknown; no other adult family member in jail in previous 5 years; total jobs in last year is 0; not homeless for 2+ days in past 5 years; reports asking God for help; male; no food stamp usage in last year.

\subsection{Debate Transcript}

\paragraph{Prosecutor 1:} (Judge Prior: No, 40\%) The defendant's history of two prior arrests within the last five years, combined with his young age and lack of employment or education, suggests a significant risk for reoffending within three years. This pattern aligns with recidivism research indicating that individuals with multiple arrests and limited prosocial factors are at heightened risk.

\paragraph{Defense 1:} (Judge Prior: Yes, 65\%) The court should consider the defendant's strong potential for rehabilitation despite his past arrests. His young age, at 22, and lack of employment or education are not necessarily predictors of future criminal behavior when viewed in isolation. Instead, we must examine his individual circumstances and positive factors that can mitigate risk.

\paragraph{Prosecutor 2:}
The defendant's history of two prior arrests within the last five years is a significant risk factor for reoffending within three years. According to the Bureau of Justice Statistics, individuals with multiple arrests have a higher likelihood of recidivism compared to those without such records.

\paragraph{Defense 2:}
While acknowledging the defendant's two prior arrests, it is crucial to consider his strong potential for rehabilitation and various positive factors that can mitigate recidivism risk. His young age of 22 suggests a significant amount of untapped growth potential. Additionally, although he lacks a high school degree or employment, his spirituality - indicated by seeking help from God - can serve as a protective factor against criminal behavior.

\paragraph{Prosecutor 3:}
The defendant's history of two prior arrests within the last five years, combined with his young age and lack of employment or education, indicates a significant risk for reoffending within three years. This aligns with recidivism research which shows that individuals with multiple arrests and limited prosocial factors have heightened recidivism rates.

\paragraph{Defense 3:}
The court should consider the defendant's strong potential for rehabilitation despite his past arrests. His young age of 22, coupled with the lack of a high school degree and employment, does not necessarily predict future criminal behavior when viewed in isolation. Instead, we must examine his individual circumstances and positive factors that can mitigate risk.

\subsection{Final Judge's Ruling}
After careful consideration of all presented arguments and evidence, it is determined that the defendant poses a significant risk for reoffending within three years. The two prior arrests within the last five years, combined with his young age and lack of employment or education, align with recidivism research indicating heightened risks for individuals with multiple arrests and limited prosocial factors. (Prediction: Yes with 70\%, INCORRECT)

\subsection{Judge Reasoning}
\begin{itemize}
    \item The defendant's history of two prior arrests within the last five years is a significant risk factor for reoffending within three years, as supported by the Bureau of Justice Statistics indicating that individuals with multiple arrests have higher recidivism rates compared to those without such records.
    \item Although the defendant's young age (22) suggests potential for growth and change, it is combined with other risk factors, such as lack of employment or education, which do not bode well for his prospects for rehabilitation.
    \item While the defendant's spirituality can serve as a protective factor against criminal behavior, this mitigating factor does not outweigh the substantial risks posed by his history of arrests and lack of prosocial factors.
\end{itemize}

\subsection{Judge Critique}
The prosecutor effectively utilized recidivism research to argue for a heightened risk of reoffending due to the defendant's multiple arrests and limited prosocial factors. Conversely, the defense attempted to highlight potential for rehabilitation based on the defendant's young age and spirituality. However, the cumulative weight of his history of two prior arrests and lack of employment or education outweighed these mitigating factors in this judge's evaluation.

\clearpage

\appendix
\section{Appendix C: Comparing Performance vs Compute (by time and tokens)}
\label{appendix:compute}

\subsection{StandardLLM: Performance vs Wall Clock Time}
\begin{figure}[!ht]  
  \includegraphics[width=\columnwidth]{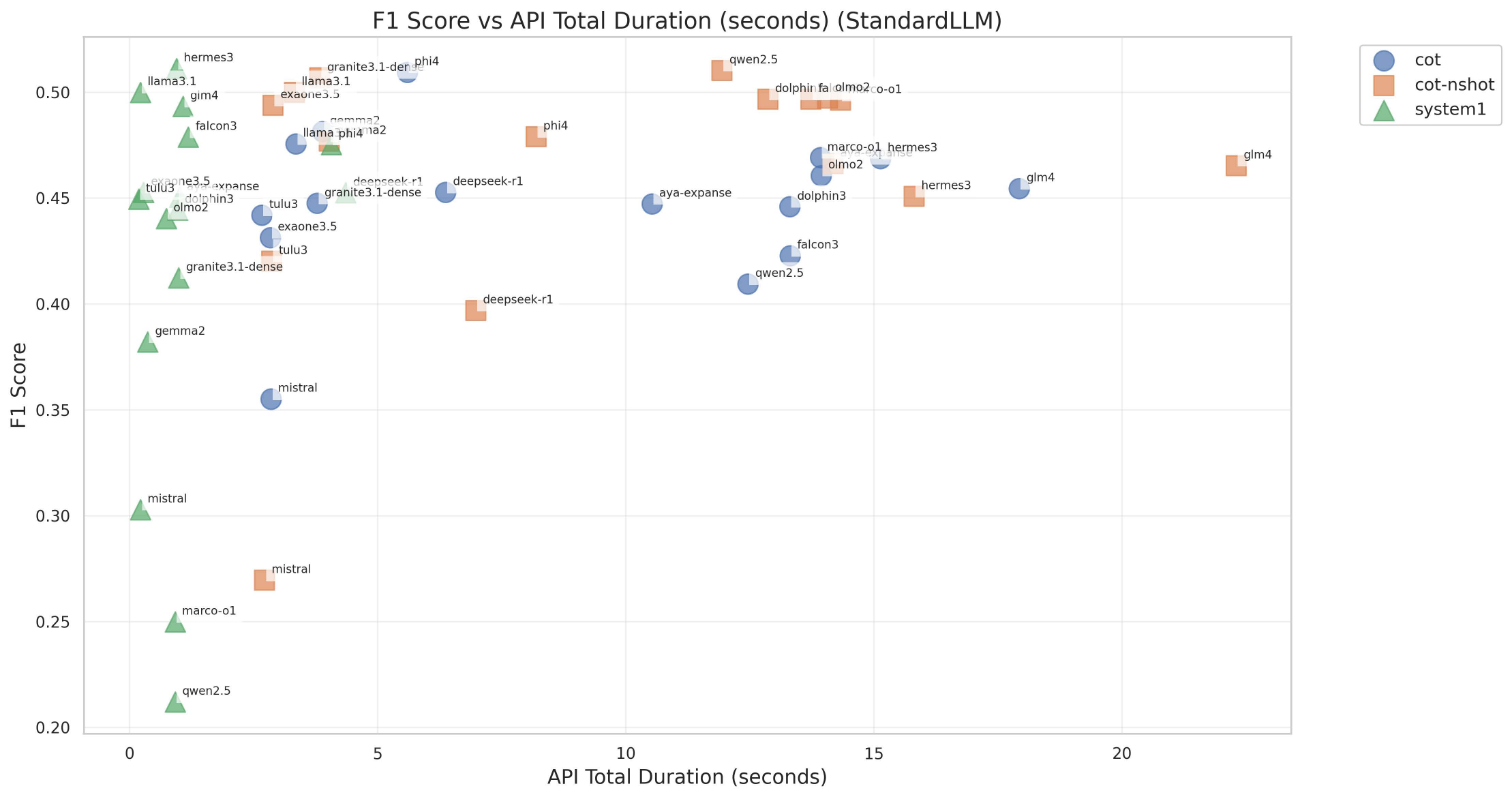}
  \caption{StandardLLM F1-score vs API Wall Clock Time)}
  \label{fig:standardllm-f1-vs-wallclock}
\end{figure}

Both token counts and API wall clock execution time were recorded for all MAD court simulations. Figure~\ref{fig:standardllm-f1-vs-wallclock} shows differences in F1-score variance by model (label) and prompt type (shape). The wider spread of system1 prompts (green triangles) and model ranking largely uncorrelation with expected performance rankings like the Huggingface Open LLM benchmark. This suggests models may be overfitting to our data on this task. Conversely, CoT prompts provide the narrowest variation in F1-scores/more stable performace albeit with top values generally below those of system1 and cot-nshot prompts. Finally, Figure~\ref{fig:standardllm-f1-vs-wallclock} show all models and prompt combination are similarly bounded by the same F1 score ceiling suggesting optimization on this task should focus on the tradeoffs between computational efficiency (system1) and generalizability (cot).

\vspace{-2mm}  

\subsection{AgenticSimLaw: Performance vs Token Count}

\begin{figure}[!hb]  
  \includegraphics[width=\columnwidth]{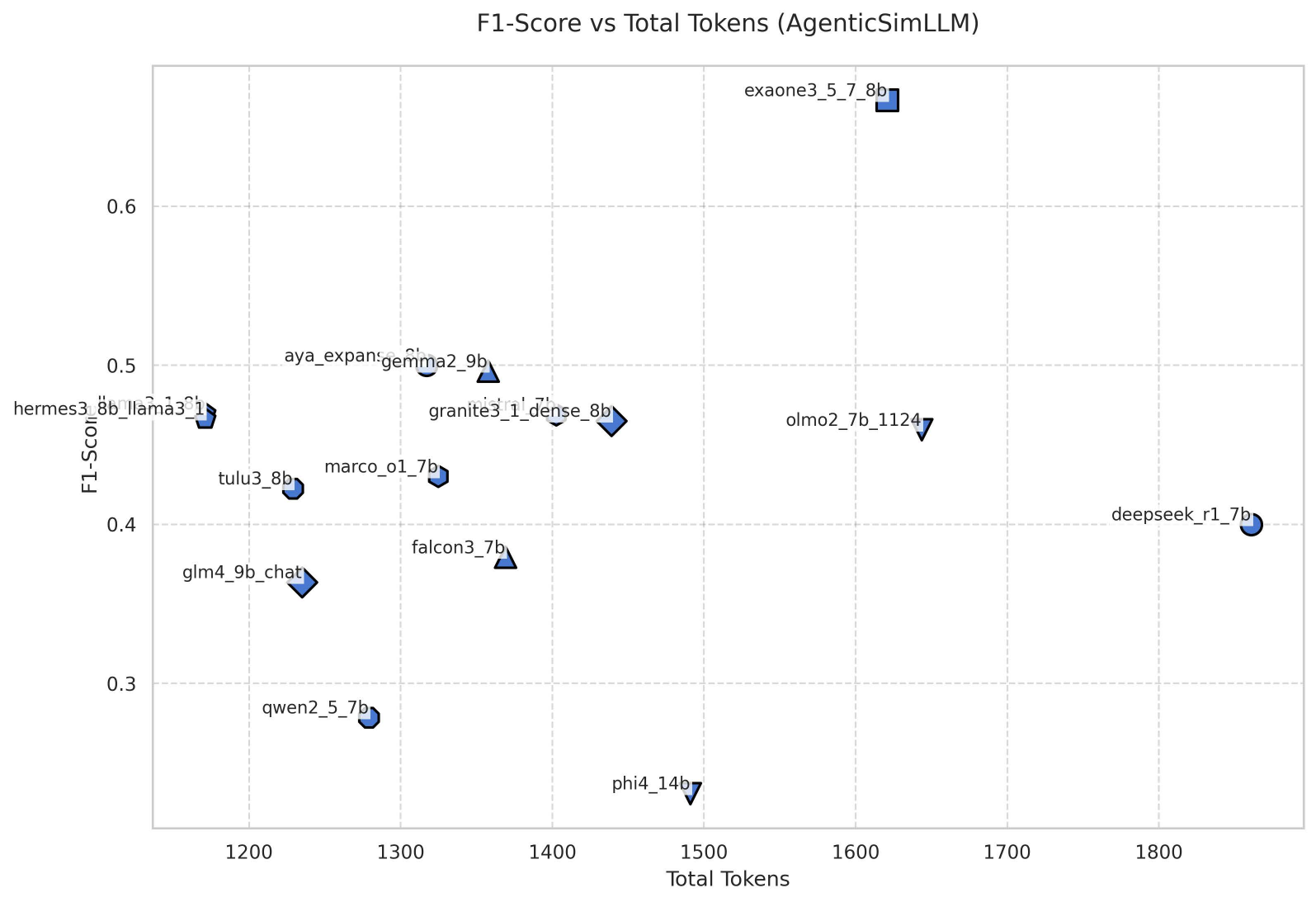}
  \caption{AgenticSimLaw F1-score vs Token Count}
  \label{fig:standard-f1-vs-tokens}
\end{figure}

Figure~\ref{fig:standard-f1-vs-tokens} focuses F1-score performance vs total token count for making recidivism predictions using MAD court simulations. Aside for the exaone 3.5 outlier, this shows the same pattern as Figure~\ref{fig:standardllm-f1-vs-wallclock}. That is, using token count as a proxy for reasoning thoroughness, we see models on the left using less reasoning has a wider range of F1 score metrics that do not correlate with popular leaderboard benchmarks. However, again we see that with more reasoning the F1 score become more stable and predictable albeit slightly below the highest scores among the band of models that do less reasoning.

\clearpage

\appendix
\section{Appendix D: MAD Simulation Stabilize Performance Metrics}  
\label{appendix:mad}

\subsection{StandardLLM: Large Ensemble}
\begin{figure}[hb!]
  \includegraphics[width=\columnwidth]{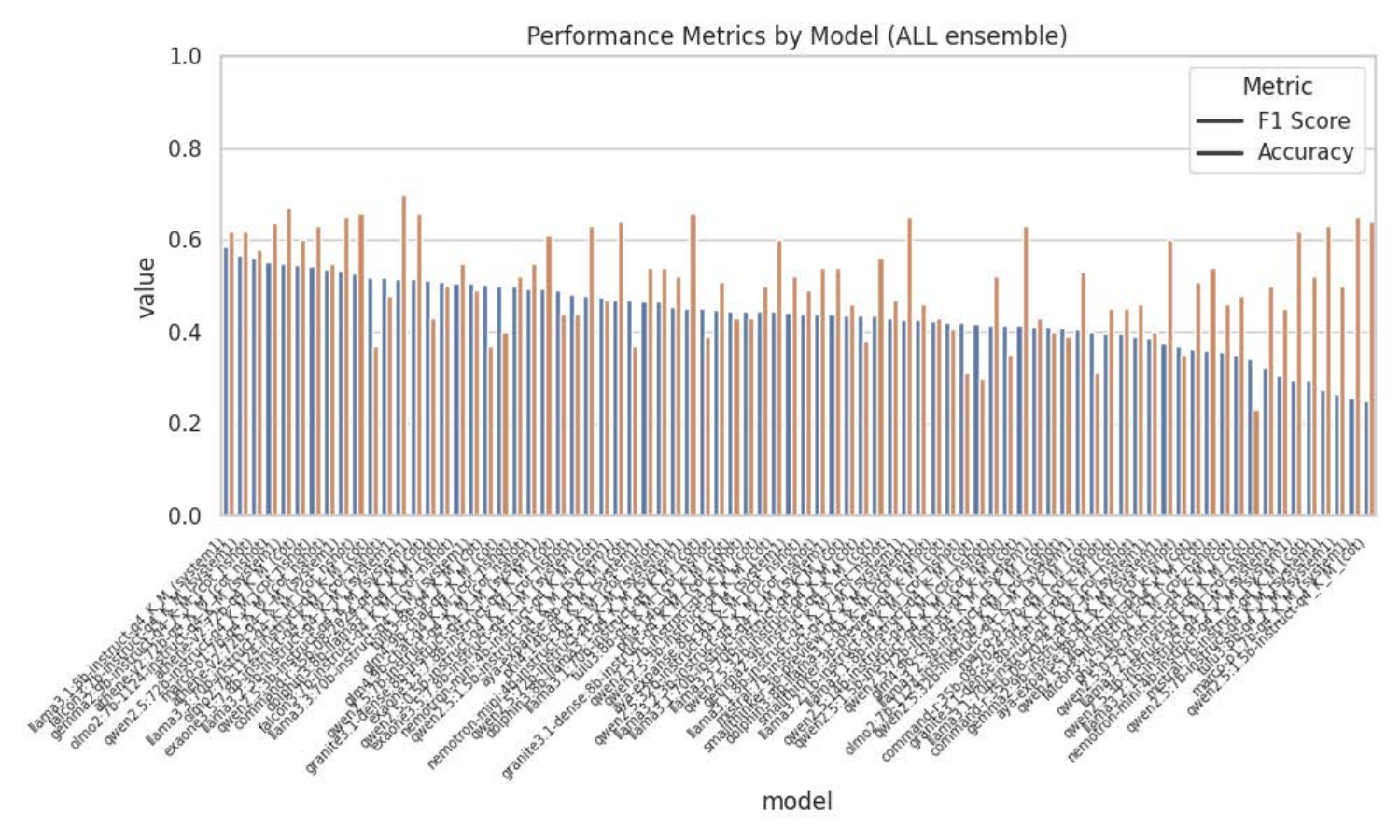}
  \caption{StandardLLM on a Large Ensemble (81 models+prompts)}
  \label{fig:standardllm_f1_accuracy_large_ensemble}
\end{figure}

Figure~\ref{fig:standardllm_f1_accuracy_large_ensemble} plots 81 unique combinations of model+prompt sorted by decreasing F1 score in blue paired with corresponding model accuracy in orange. Note the top F1 scores, are dominated by a concentration of large parameter models using cot-nshot (e.g. athene 72b, qwen 2.5 72b, and llama 3.3 70b) and unexpected smaller models (e.g. llama 3.1 8b, falcon3 7b, llama 3.2 3b). The systematic concentration of large models contrasts with the almost randomize order of smaller models. This again suggests smaller models are overfitting to the data while the large models may be more generalizable. 

\subsection{StandardLLM: Small Ensemble}
\begin{figure}[hb!]
  \includegraphics[width=\columnwidth]{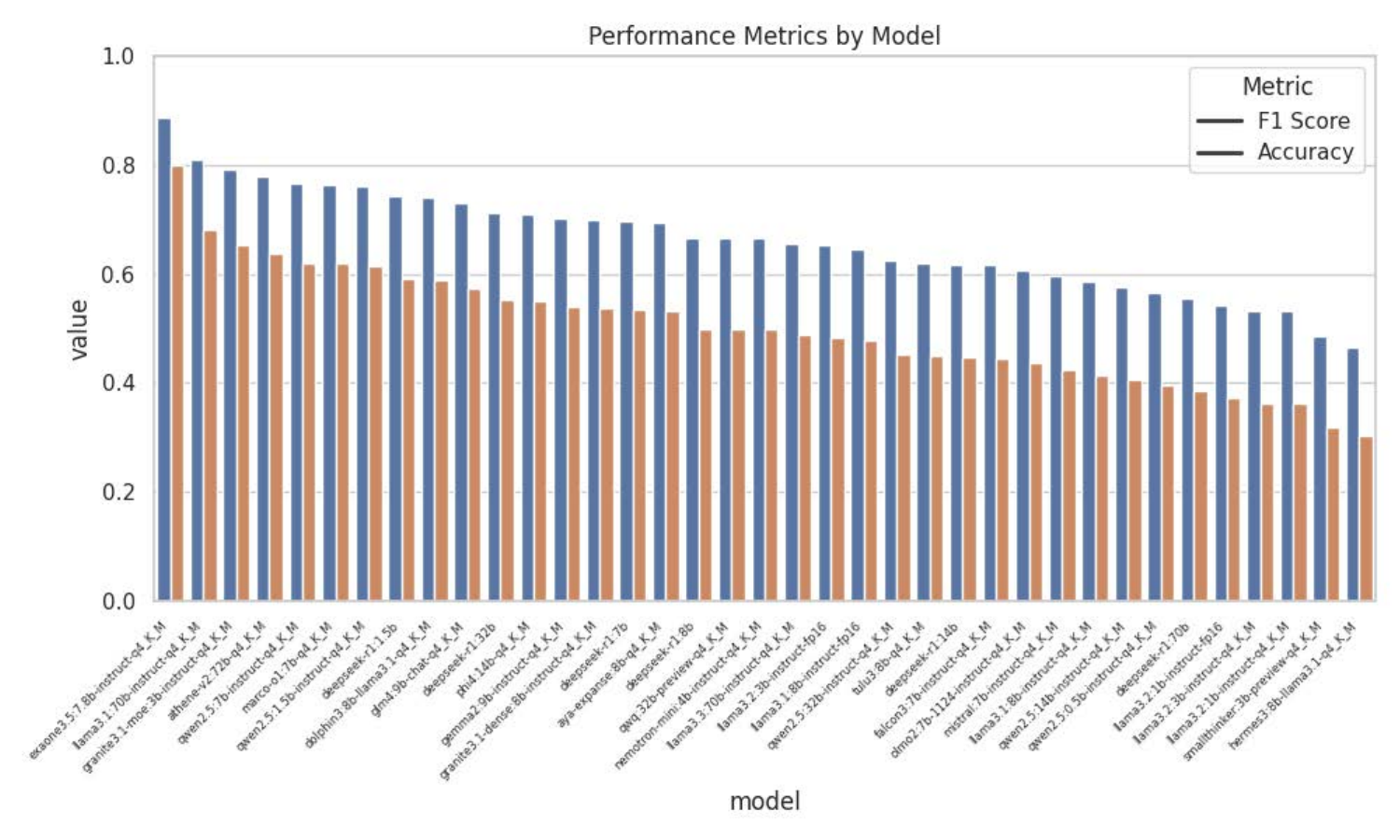}
  \caption{AgenticSimLaw on a Medium Ensemble (37 models+prompts)}
  \label{fig:agenticsimlaw_accuracy_medium_ensemble}
\end{figure}

In contrast to the StandardLLM large ensemble in Figure~\ref{fig:standardllm_f1_accuracy_large_ensemble}, the medium ensemble using AgenticSimLaw in Figure~\ref{fig:agenticsimlaw_accuracy_medium_ensemble} highlights several points. First, the F1-score is highly correlated with accuracy in a range (0.47-0.87) consistently above the range for StandardLLM's F1-score (0.09-0.58). This suggests that agentic thinking regularizes performance and makes small model accuracy a much more reliable and generalizable metric. Second, two of the three large +70b LLMs rank in the top four confirming larger models more reliably benefity from agentic reasoning on this task. Third, among the highly-regarded recent reasoning models the performace is poor to mediocre ( Deepseek-r1, Tulu3, OLMo2) suggesting internally reasoning models offer no additional advantages over our explicit external structured reasoning using MAD court simluations \citep{deepseek:25,lambert:25,teamolmo:25}. Finally, the Llama models illustrate how performance roughly increases with both model size and more quantization levels (fp16 > 4 bit) under our AgenticSimLaw reasoning method.

\clearpage

\clearpage
\appendix
\section{Appendix E: SOTA Linguistic vV Statistical Reasoning}  
\label{appendix:sota_llm_results}

Although this paper is not focused on optimizing performance metrics for our task, readers may be interested in this related topic. Here we provide SOTA performance metrics on tabular data tasks for (a) traditional statistical machine learning models (ML models), (b) specialized tabular LLMs, and (c) current leading SOTA general LLMs on our recidivism prediction task. 

\subsection{General Purpose LLM}

\begin{table}[htbp]
\centering
\begin{tabular}{llcccccc}
\hline
\textbf{Model} & \textbf{Prompt} & \textbf{Accuracy} & \textbf{Precision} & \textbf{TP} & \textbf{TN} & \textbf{FP} & \textbf{FN} \\
\hline
OpenAI gpt-4o-mini & system1 & 0.48 & 0.60 & 29 & 10 & 19 & 42 \\
 & cot & 0.71 & 1.00 & 71 & 29 & 0 & 0 \\
 & cot-nshot & 0.47 & 0.49 & 23 & 5 & 24 & 48 \\
\hline
OpenAI o3-mini & system1 & 0.49 & 0.63 & 31 & 11 & 18 & 40 \\
 & cot & 0.53 & 0.60 & 32 & 8 & 21 & 39 \\
 & cot-nshot & 0.70 & 0.96 & 67 & 26 & 3 & 4 \\
\hline
Anthropic Claude haiku-3-5 & system1 & 0.34 & 0.21 & 7 & 2 & 27 & 64 \\
 & cot & 0.71 & 1.00 & 71 & 29 & 0 & 0 \\
 & cot-nshot & 0.71 & 1.00 & 71 & 29 & 0 & 0 \\
\hline
Anthropic Claude sonnet-3.5 & system1 & 0.43 & 0.44 & 19 & 5 & 24 & 52 \\
 & cot & 0.53 & 0.68 & 36 & 12 & 17 & 35 \\
 & cot-nshot & 0.57 & 0.72 & 41 & 13 & 16 & 30 \\
\hline
\end{tabular}
\caption{Performance using SOTA Commercial LLMs}
\label{tab:sota-generalllm_performance}
\end{table}

\subsection{Statistical ML and Specialized Tabular LLMs}
Table~\ref{tab:sota-all_performance} compares the median performance metrics for both the leading traditional statistical ML model (XGBoost) and two popular specialized tabular LLMs (TabPFN, TabNet) \citep{ma:24}. These metrics are based on performance over 48 OpenML tabular datasets where ICD boosts performance with "in-context distillation" (ICD) for more efficent use of n-shot examples. The recent tabular LLM TabPFN model claims SOTA peformance in low data regimes with datasets under 10,000 samples \citep{hollmann:25}. Note, performance metrics for these models on our recidivism prediction tasks are likely to differ from these benchmark OpenML datasets.

\begin{table}[h]
\centering
\label{tab:model-performance}
\begin{tabular}{lccc}
\hline
\textbf{Model} & \textbf{Median AUC} & \textbf{Median F1} & \textbf{Median Accuracy} \\
\hline
XGBoost (Tuned) & 0.969 & 0.921 & 0.923 \\
TabPFN-ICD & 0.967 & 0.899 & 0.902 \\
XGBoost & 0.953 & 0.893 & 0.894 \\
TabPFN & 0.951 & 0.847 & 0.844 \\
TabNet & 0.939 & 0.887 & 0.887 \\
\hline
\end{tabular}
\caption{SOTA Metrics for Statistical ML and Specialized Tabular LLMs}
\label{tab:sota-all_performance}
\end{table}

\end{document}